\documentclass[letterpaper, 10 pt, conference]{ieeeconf}  % Comment this line out if you need a4paper

\IEEEoverridecommandlockouts                              % This command is only needed if 
\usepackage{multicol}
\usepackage[bookmarks=true]{hyperref}
\usepackage{multirow}
\usepackage{times}
\usepackage{epsfig}
\usepackage{graphicx}
\usepackage{amsmath}
\usepackage{amssymb}
\makeatletter
\let\NAT@parse\undefined
\makeatother
\usepackage[numbers]{natbib}

\title{\LARGE \bf
Learning an Action-Conditional Model for Haptic Texture Generation
}

\author{Negin Heravi, Wenzhen Yuan, Allison M. Okamura, and Jeannette Bohg 
\thanks{
\newline N.\ Heravi and A.~M.\ Okamura are with the Department of Mechanical Engineering, Stanford University. W.\ Yuan is with the Robotics Institute at Carnegie Mellon University. J.\ Bohg is with the Department of Computer Science, Stanford University.{\tt\small [nheravi,aokamura,bohg]@stanford.edu, wenzheny@andrew.cmu.edu}. 
\newline N. Heravi was supported by the NSF Graduate Research Fellowship. This work has been partially supported by Amazon.com, Inc. through an Amazon Research Award. This article solely reflects the opinions and conclusions of its authors and not of Amazon or any entity associated with Amazon.com. Research reported in this publication was also partially supported by the 2019 Seed Grant from the Stanford Institute for the Human-Centered Artificial Intelligence (HAI).
\newline We thank Katherine Kuchenbecker and Yasemin Vardar for giving us access to the textures used in the Penn Haptic Texture Toolkit (HaTT), Shaoxiong Wang for the GelSight sensor, and Heather Culbertson for answering our questions regarding HaTT.
}}

\begin{document}

\maketitle
\thispagestyle{empty}
\pagestyle{empty}

%%%%%%%%%%%%%%%%%%%%%%%%%%%%%%%%%%%%%%%%%%%%%%%%%%%%%%%%%%%%%%%%%%%%%%%%%%%%%%%%

\begin{abstract}
%A virtual reality or teleoperation system is most effective if it provides a truly immersive experience to the user. 
Rich haptic sensory feedback in response to user interactions is desirable for an effective, immersive virtual reality or teleoperation system. However, this feedback depends on material properties and user interactions in a complex, non-linear manner. Therefore, it is challenging to model the mapping from material and user interactions to haptic feedback in a way that generalizes over many variations of the user's input. Current methodologies are typically conditioned on user interactions, but require a separate model for each material. 
In this paper, we present a learned action-conditional model that uses data from a vision-based tactile sensor (GelSight) and user's action as input. This model predicts an induced acceleration that could be used to provide haptic vibration feedback to a user. We trained our proposed model on a publicly available dataset (Penn Haptic Texture Toolkit) that we augmented with GelSight measurements of the different materials. We show that a unified model over all materials outperforms previous methods and generalizes to new actions and new instances of the material categories in the dataset.
\end{abstract}

%%%%%%%%%%%%%%%%%%%%%%%%%%%%%%%%%%%%%%%%%%%%%%%%%%%%%%%%%%%%%%%%%%%%%%%%%%%%%%%%
\section{Introduction}
\label{sec:Introduction}
% With the commercialization of several virtual reality devices over the past few decades \jean{citation}, a diverse variety of applications in several areas such as e-commerce, gaming, education, and healthcare has emerged. However,

Realistic virtual reality (VR) environments benefit from rich multi-modal sensory feedback, including the visual, haptic, and auditory signals that humans normally receive during real-life manipulation tasks. Humans are conditioned to expect the sense of weight, hardness, deformability, texture, and slipperiness when interacting with an object \cite{NoeBook,BelievabilityinVR,KlatzkyTextureProbe,DBLP:journals/corr/BohgHSBKSS16}, an experience that does not fully exist in commercially available VR systems. To this end, several researchers have rendered textures by varying the magnitude and direction of the force imposed on the user using a force feedback device \cite{Costa2000RoughnessPO,AllisonHoweTexture,Culbertson2014ModelingAR}, by varying local surface friction using a surface haptic display \cite{T-pad}, or by using a voice coil motor to induce controlled acceleration signals in a hand-held pen \cite{Culbertson2014OneHD, Romano}. These approaches develop a separate model for each texture, which makes it hard to scale them to the unlimited variety of textures in the world. 
There are approaches that learn a joint latent representation for different textures who show generalization to novel inputs. However, they focus
on texture classification or property estimation and not on generating haptic feedback \cite{ConstrainedLMT,LMTMulti, Burka,Takahashi}.

%Automated data-driven methods for modeling these sensations are useful as robots can collect data that would then be modeled and displayed in the virtual environment. 

In this paper, we focus on data-driven modeling of the vibratory feedback that different textures induce in a probe as it is moved over a surface. This vibration is linked to humans' perceptual impression of texture and is a function of the probe's action as well as the texture \cite{KlatzkyTextureProbe}. This form of haptics was selected for study because of the availability of public datasets in this area. 
We propose a novel learning-based method for haptic texture generation using an action-conditional model. This model takes as input (i) an image from the GelSight tactile sensor~\citep{GelSight1,GelSight2} while pressed on a texture, and (ii) force and speed of the user on that texture during a tool-mediated interaction over a horizon of 1 ms. Given this input, we train a model that predicts the magnitude of the discrete fast Fourier transform of the generated acceleration in the hand-held probe within the next 0.1 s. We predict the spectral content instead of the temporal signal because there is evidence that human texture sensation is invariant to phase shifts \cite{phasenotmatter}.

We train our model using supervised learning data from the {\em Penn Haptic Texture Toolkit\/} (HaTT) \cite{Culbertson2014OneHD}. We show that our novel methodology for generating haptic textures learns a unified model for several textures and generalizes to new force and speed interactions. The learned latent representation of different textures places materials that feel similar closer to each other. This opens the opportunity to generate haptic textures of new materials by locating them within that latent space. In sum, this paper presents a new action-conditional model for generating haptic textures that is learned from data. The primary contributions are as follows:
\begin{enumerate}
\item  By concatenating human actions with a feature representation of a GelSight image of the texture, our model can predict the vibratory feedback in a hand-held probe during user interactions.
\item Our proposed model is unified across different textures, reducing the need for developing a separate model for each texture instance. 
\item Our model generalizes to previously unseen force and speed interactions as well as new instances of the modeled textures and outperforms prior work in terms of DFT prediction accuracy. 
\end{enumerate}
Additionally, we have augmented the HaTT dataset with GelSight videos available at: \href{https://sites.google.com/stanford.edu/haptic-texture-generation}{sites.google.com/stanford.edu/haptic-texture-generation}

\begin{figure*}[ht]
\begin{center}
%\fbox{\rule{0pt}{2in} \rule{0.9\linewidth}{0pt}}
\includegraphics[width=.95\linewidth]{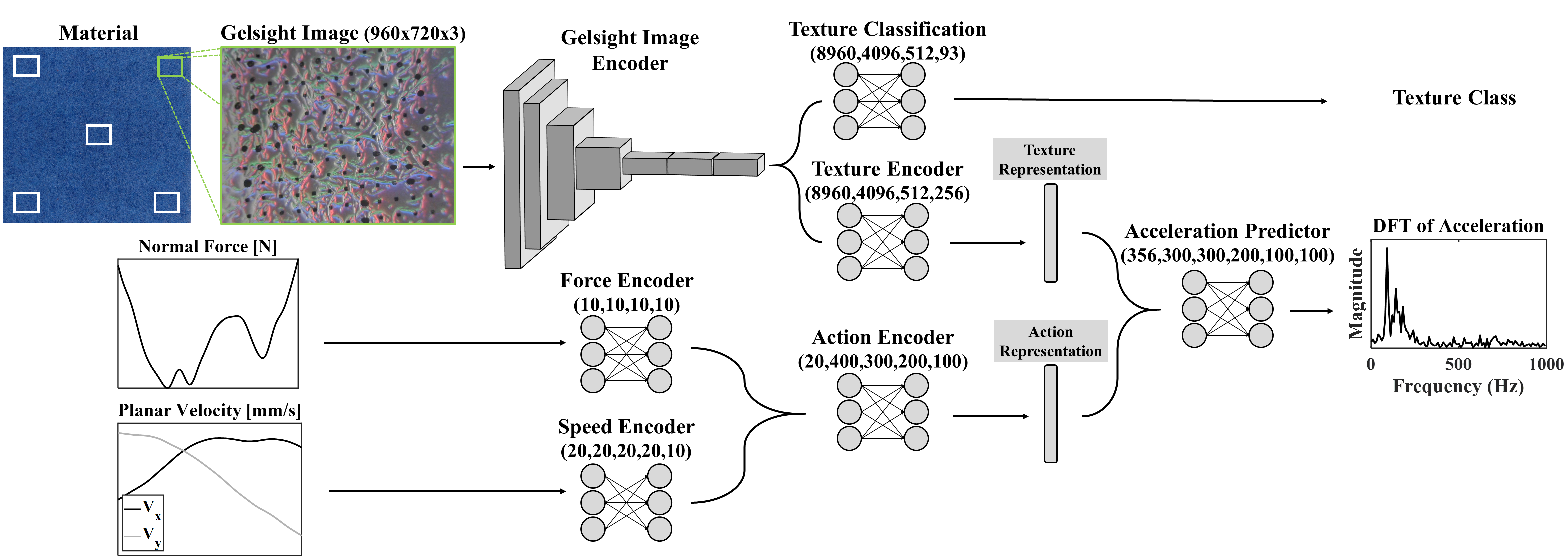}
\end{center}
\caption{Neural Network architecture for short horizon haptic signal generation. This model takes a GelSight image of the material as well as the force and speed imposed by the user over a 1 ms window on the material surface (action) and predicts the magnitude of the spectral content of acceleration in the next 100 ms induced in the material due to this interaction.}

\label{fig:Architecture}
\end{figure*}

\section{Related Work and Background}
\label{sec:Background}

Materials respond to user interactions in a highly variable and non-linear fashion. This makes manual modeling of haptic feedback hard. Therefore, researchers have explored data-driven approaches. For data-driven voice-coil based texture rendering, there are approaches based on linear predictive coding (LPC) \cite{Romano}, autoregressive moving average (ARMA) \cite{RefinedARMA} and piece-wise autoregressive moving average \cite{Culbertson2014OneHD}. However, these approaches are challenging to scale to the unlimited number of textures around us because they learn a separate model for each texture and do not relate them to each other in a joint model. Thus, as the number of desired textures for rendering increases, the number of saved models linearly increases. Furthermore, this set of models cannot generalize to new textures as it is unclear how this new texture is similar or different from the textures instances in the training set.

% For example, there is no connection between the models generated for two different sandpapers even though they perceptually feel close to each other.
There is work that learns a joint model of different textures which generalizes to novel textures. \cite{ConstrainedLMT,LMTMulti, Burka} focus on using data collected during tool-mediated interactions to extract surface properties. As input, the proposed models use a variety of different modalities such as recorded acceleration, sound, normal force, friction, and RGB images -- either individually or in combination. The output of the models are properties that are then used as a texture representation feature which helps to evaluate the similarity of different textures for texture classification. However, \cite{ConstrainedLMT,LMTMulti, Burka} do not use this representation for generating haptic feedback. More recently, \citet{Takahashi} have provided evidence that a network can be trained to estimate tactile properties of new surfaces using an RGB image as input and shown the value of a learned latent representation vector in estimating tactile properties, but the performance of their model on predicting temporal data is unclear.

Deep learning-based approaches can learn complex functions which are hard to model manually. They learn feature representations from data and can generalize to new data \cite{Michelle,Wenzhen}. Inspired by this, we propose a learning-based approach that uses the image of a texture and the action of the user to generate haptic texture feedback. In our approach, we learn a joint model over many textures such that it can generalize to novel textures. Furthermore, the experienced sensation of a texture is not just a function of the material but also the imposed force and speed on that texture by the user. Therefore, we propose an action-conditional model.

%\subsection{For Negin Note to self}
%a quantitative measure of generalization performance of such %models remains unknown. 

%The capability of require a predetermined separate model of a surface before it can be rendered to the users limiting the available options. For such models to be generated for a new texture, a costly data collection process using a variety of expensive sensors and equipment is required. The cost of this process can be significantly mitigated if one could use the visual properties of these textures to predict their haptics feeling without further data collection.

\section{Generative Action-Conditional Model}
\label{sec:GenerativeModel}

The inputs of this structure are a GelSight image, obtained by pressing the sensor on the material, as well as a user's force and speed measurements during 1 ms (10 data points) of interaction with the texture. Our model outputs the magnitude of the discrete Fourier transform (DFT) of the induced acceleration for the next 100 ms. This acceleration corresponds to human's perceptual impression of texture. The goal of this model is to infer the relationship between a combination of a human's action and texture's representation with the corresponding induced acceleration. 
%During a live demo, this model could be used in the following example application. First, we collect a GelSight image of the material that would be encoded in a texture representation vector. Afterwards, the user's force and speed are recorded as they move their hand in virtual reality. These readings are fed into our model to predict the magnitude of the short term DFT of the expected acceleration. This short term spectral prediction is then used to construct a temporal acceleration signal that can be displayed to the user through vibratory feedback in real time. 

\subsection{Neural Network Architecture}
Fig.~\ref{fig:Architecture} shows the architecture that enables the short-term DFT prediction of our model. This architecture encodes the GelSight image into a texture representation vector and combines it with the encoded action representation from the user's force and speed to predict the desired DFT using an acceleration predictor module. This structure was chosen empirically through a hyper-parameter search.

We use AlexNet \cite{Alexnet} with fine-tuned weights as image encoder with two additional CNN layers to further decrease the resolution of the input images (960$\times$720$\times$3). This encoder was chosen based on its optimal performance on the proxy task of classifying the GelSight images. The other encoders or layers in the model are fully connected (FC) with rectified linear layers in between.
 
The architecture is trained in two stages. First, we train the image encoder augmented with three fully connected layers for texture classification using a cross-entropy loss. 
Afterwards, we freeze these pre-trained weights and use the output of the image encoder as input to the texture encoder. We then train the full architecture for predicting the DFT magnitude of the accelerations. The choice of freezing over fine-tuning was motivated by its better performance on the validation set. As loss we choose the Euclidean distance between the ground truth and predicted magnitude up to 1000~Hz (100 DFT bins). 

 %speed encoder = (20,20,20,20,10), force encoder = (10,10,10,10), classification architecture = (8960,4096,512,93), 
%texture encoder = (8960,4096,512,256), action encoder = (400,300,200,100), acceleration predictor = (356,300,300,200,100)

%\begin{figure}[ht]
%\begin{center}
%\fbox{\rule{0pt}{2in} \rule{0.9\linewidth}{0pt}}
%\includegraphics[width=1\linewidth]{fig/Classification.png}
%\end{center}
 %  \caption{Training and validation plots for the classification loss for four different models: Model 1: Fine-tuning a pre-trained Alexnet architecture with no augmentation on the dataset, Model 2: Fine-tuning a pre-trained Alexnet architecture with dataset augmentation, Model 3: Model 2 with an image size of (480x360) as an input. This model had two additional layers of CNN on top of the Alexnet feature extractor architecture to decrease the dimensionality of the output. Model 4: Model 3 with an image size of (960x720) as an input.  Based on these results the architecture from the feature extractor in the fourth model and its tuned weights were chosen for the initialization of the haptic rendering model. 
%}
%\wenzhen{Would it be better to make this figure in the Experiment section? I would also put this figure in a low priority if you must remove some figures from the paper.}
%\label{fig:Classification}
%\end{figure}

\subsection{Temporal Signal Construction}
\label{HapticRendering}

Rendering haptic textures requires a long-term temporal vibratory signal. To construct such signal from the predicted short-term DFT magnitude of our model, we use \citet{Prusa2017ThePR}'s implementation of the Griffin and Lim Algorithm (GLA) \cite{GLA} in Matlab. To achieve faster convergence, we use a variation of GLA called fast GLA \cite{FGLA}. The caveat of using GLA is that it does not run in real time. However, similar optimization-based online phase retrieval and signal reconstruction algorithms that are capable of running online exist and will be explored in future work \cite{Prusa2017ThePR}. 

%To provide a basic proof of concept that our method can work in real-time, we also directly stitched the inverse Fourier transformed sequence (with random phase) and report the performance using this constructed long term temporal prediction in the experimental section. 

\section{Dataset}

%\begin{figure}[ht]
%\begin{center}
%\includegraphics[width=1\linewidth]{fig/RGB_image.png}
%\end{center}
%   \caption{(a) RGB Image of scouring pad. White rectangles indicate the approximate 5 locations at which GelSight videos were collected (b) A sample corresponding GelSight image}
%   \wenzhen{For showing the example of GelSight data, I think (a) is not that necessary. Instead, it would be nice if you show 1) multiple GelSight data on one material, and 2) GelSight data on multiple materials.}
%\label{fig:RGBGel}
%\end{figure}

Researchers in the computer vision community have studied and published purely image-based texture databases %such as Brodatz repository~\cite{Brodatz}, VisTex~\cite{VisTex}, Meastex ~\cite{MeasTex}, and Dyntex~\cite{DynTex} 
for several years \cite{Brodatz,VisTex,MeasTex,DynTex}. However, these datasets lack multi-modal sensory information such as haptic information. Only a few limited databases have been made publicly available in the haptics community. Specifically, the Penn Haptic Texture Toolkit (HaTT)~\cite{Culbertson2014OneHD} and the LMT Haptic Texture Database~\cite{ConstrainedLMT,LMTMulti} are the two main publicly available haptic texture databases. The sensory data collected in the LMT haptic is suitable for texture classification but is unsuitable for our task as they lack positional tracking of the tool. Hence, our model was trained using HaTT~\cite{Culbertson2014OneHD}.

HaTT includes raw data used to create haptic texture and friction models for one hundred different materials from diverse categories including paper, metal, carbon fiber, fabric, plastic, wood, stone, foam, tile, and carpet. The original data was collected by a sensorized pen providing 6 DoF force/torque readings, acceleration measurement, as well as positional and orientational tracking of the pen's top. For each material, \citet{Culbertson2014OneHD} used this pen to measure a 10-second signal of a human's unconstrained circular motion. % The dataset uses the position and orientation data from the tracker to provide the speed of the pen's top using discrete-time derivatives. Furthermore, it uses the known orientation of the pen to calculate the normal and tangential force to the surface. A variety of processing methods were used to mitigate noise in the data and remove auditory frequencies that may interfere with human's perception described in detail in \cite{Culbertson2014OneHD}.
The resulting dataset (HaTT) includes textural haptic information about a large variety of textures making it a suitable choice for evaluating our model. %\wenzhen{These four sentences could be removed}However, using this dataset also introduces a challenge: one image per texture is provided in this dataset making it unsuitable for deep learning based data-driven modeling. \wenzhen{I actually don't think you need to mention how the image is not helpful, and this sentence is not also very well justified... You can just say that you went ahead to collect GelSight, because GelSight data contains most important modeling info about the material} To address this difficulty, 

We have augmented this dataset with GelSight images. GelSight~\cite{GelSight1,GelSight2} is a vision-based high resolution tactile sensor made of a piece of clear elastomer coated with a reflecting membrane. A video-recording camera is attached to the other side of this elastomer and captures its deformation during contact. These deformations provide high resolution information about the geometry of the surface they are in contact with. Using this sensor, we have collected videos of 5 presses on 93 out of the 100 materials due to availability during data collection. The location of these presses (shown by white rectangles) as well as an example GelSight image are shown as the input of the image encoder in Fig. \ref{fig:Architecture}. 
This addition provides the opportunity to explore research questions on the advantages of using GelSight over RGB in texture rendering or hardness estimation using a similar approach to that of \citet{Yuan2017ShapeindependentHE} in future work.

%\begin{figure}[ht]
%\begin{center}
%\includegraphics[width=1\linewidth]{fig/RGB_image.png}
%\end{center}
   %\caption{Left image: RGB Image of scouring pad. White %rectangles indicate the approximate 5 locations at %which GelSight videos were collected. Right image: a %sample corresponding GelSight image}
%\label{fig:RGBGel}
%\end{figure}

\begin{figure}[t!]
\begin{center}
%\fbox{\rule{0pt}{2in} \rule{0.9\linewidth}{0pt}}
\includegraphics[width=.955\linewidth]{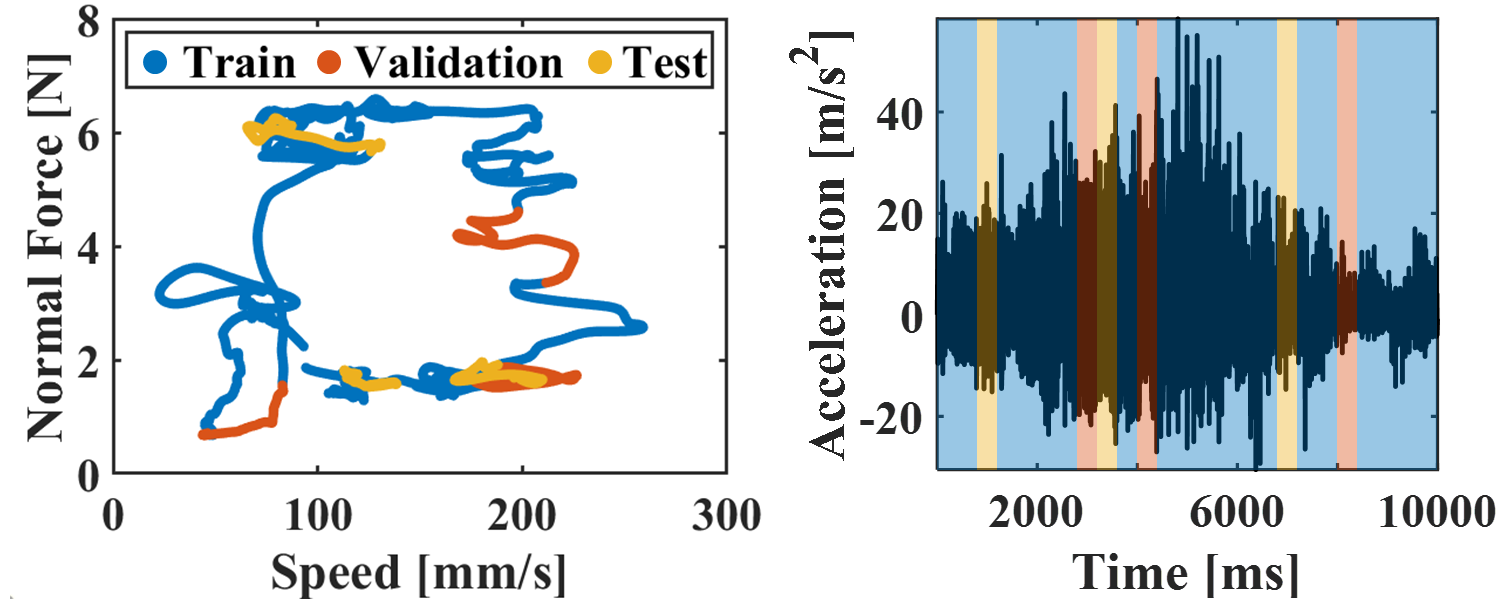}
\end{center}
   \caption{Train, validation and test set divisions for ABS plastic: (left) shows how this division corresponds to different data points in force and speed space and (right) shows the same division on top of the acceleration signal in temporal domain}

\label{fig:TrainValDiv} 
\end{figure}
 
\subsection{Data Preprocessing}

%\begin{figure}[t]
%\begin{center}
%\fbox{\rule{0pt}{2in} \rule{0.9\linewidth}{0pt}}
%\includegraphics[width=1.0\linewidth]{fig/All2Force.jpeg}
%\end{center}
%   \caption{Example Normal Force measurements from the dataset for two different materials. It can be observed that the high frequency part of the force measurement is related to the material textures. Consequently, to only capture the user's action low-passed filtered force and speed were used as the input to our model.}
 %  \negin{to do: (minor)change the legend so it's more readable}
%   force is not simply a function of user's action but also contains information regarding the texture and needs to be low pass filtered to extract the action information. 

%\label{fig:Lowpass}
%\end{figure}

Force and speed measurements during interactions do not merely include the action of the user, but also oscillations generated by the texture. To mitigate the texture's influence, we low-pass filtered the force and speed signals at 20 Hz before feeding them into the neural network. 
%Example forces for two materials and their low-pass filtered signal is shown in Fig. \ref{fig:Lowpass}.
 
% Furthermore, to be able to test the generalizability of our method on new actions, w
To build the training, validation, and test sets, we divide each interaction sequence into 25 sections and re-group them. An example is shown in Fig. \ref{fig:TrainValDiv}. 
% We divide each interaction sequence into training, validation, and test sets by cutting it into 25 chunks, and an example is shown in Fig. \ref{fig:TrainValDiv}. 
The test and validation sets are then chosen such that their
%bounding boxes in the 
force and speed regions
%space has a non zero 
overlap with those of the training set. This was imposed to increase the likelihood of the test and training data sharing similar sampling distributions. 
Out of the 5 collected GelSight videos (each containing a single press), 3 were used for training, 1 for validation, and 1 for testing. These videos were processed by extracting the frame that has the highest pixel value difference from the non-contact frame. This frame usually corresponds to when the sensor is making the largest contact force with the material. For the videos used in the training set, the adjacent frames of this peak frame were also used for a total of 12 images per video. For further augmentation, these images were rotated at multiples of 90 degrees and mirrored.

\section{Experiments}
\label{sec:Experiments}
%\wenzhen{I think somewhere in this paper, you might need to explicitly explain where the 'representation' comes from, and why. My understanding is that, the representation comes from only the GelSight image, regardless of the motion (force and speed), because the GelSight images are expected to sample and describe the material texture. When touching different parts of one material with GelSight, you have different 'observation', but the material should be the same, so the representation is the same. And the output vibration data is a result of both the material and the action. For Heather's method, they also have a representation of the material, which is the coefficients of the AR model. They learnt the representation from the historical force&speed data, not some other observation. } I referred to the architecture what I exactly mean by representation vector. It's a bit hard to explain as if we make a claim about it being a specific thing we need to prove which we haven't but it's more than just image as it also places material that feel and not just look similar close . 

The primary goal of our experiments is to evaluate the performance of our 
% action-conditional neural-network-based 
model for the purpose of haptic texture acceleration signal generation. The experiments are driven by three main questions:
\begin{enumerate}
    \item Can this model generate the induced haptic acceleration signal of a texture given the user's actions? 
    \item Is there an advantage to learning a unified model for different textures based on their GelSight image as opposed to modeling each material separately?
    \item Does the model \textit{generalize} to: (i) new interactions by the user, (ii) new GelSight images of training materials, and (iii) GelSight images of novel materials?
\end{enumerate}

In all of our experiments, we conducted training on a Quadro P5000 GPU using Pytorch \cite{Pytorch} with Adam as the optimization algorithm \cite{Adam}. The t-SNE plots were generated using the implementation in Scikit-learn \cite{TSNE}. 

\begin{figure*}[t]
\begin{center}
%\fbox{\rule{0pt}{2in} \rule{0.9\linewidth}{0pt}},angle=90
\includegraphics[width=0.99\linewidth]{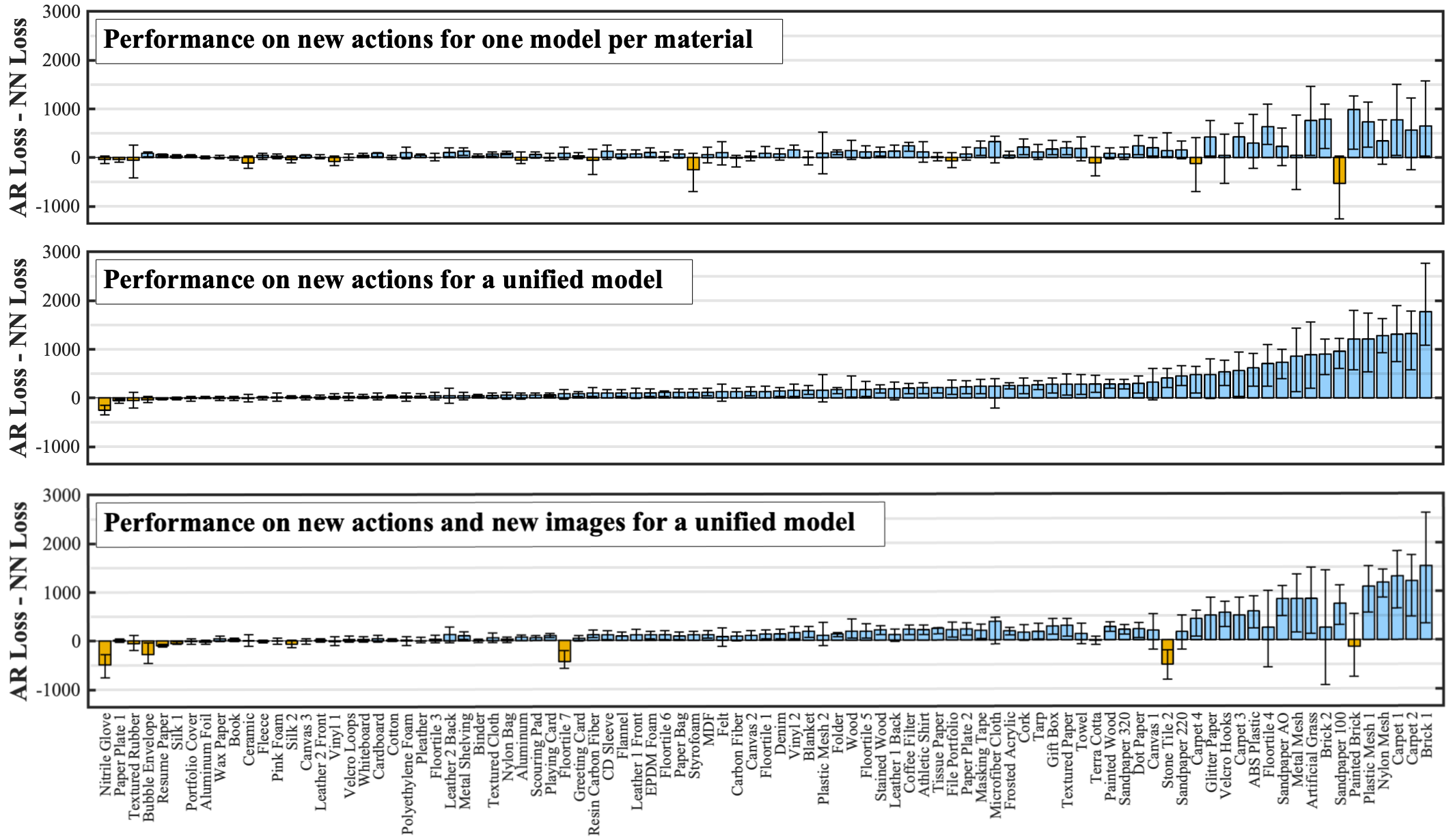}
   \caption{Comparison of the AR model with the NN based model for (top) generalization to new actions using one model per material, (middle) generalization to new actions using unified model for all materials and (bottom) generalization to new actions and GelSight images using unified model for all materials. This histogram represents the difference between the loss of the AR prediction with that of a NN prediction. Error bars indicate 25 and 75 percent quartiles. The loss for each model is defined as the Euclidean distance between the local short term DFT of its prediction and the corresponding spectral content of the ground truth acceleration. A positive value in the bar plot means our method outperforms the AR model.}
\label{fig:Shortterm}
\end{center}
\end{figure*}

%\begin{figure}[t]
%\begin{center}
%\fbox{\rule{0pt}{2in} \rule{0.9\linewidth}{0pt}},angle=90
%\includegraphics[width=0.5\textwidth]{fig/BlueBars_vertical.png}
%   \caption{Comparison of the AR model with the NN based model for (top) generalization to new actions using one model per material,(middle) generalization to new actions using unified model for all materials and (bottom) generalization to new actions and GelSight images using unified model for all materials. This histogram represents the difference between the loss of  the AR prediction with that of a NN prediction. The error bars indicate the 25 and 75 percent quartiles. The loss for each model is defined as the Euclidean distance between the local short term DFT of its prediction and the corresponding spectral content of the ground truth acceleration. A positive value in the bar plot means our method outperforms AR model. \allison{ did you try making it a vertical set plots? I'm curious what that looks like.}}
%   \negin{I did but can we please keep it horizion. I really dislike the vertical version as I can't read anything when I print it out, but to see that version uncomment the following figure}
%\label{fig:Shortterm}
%\end{center}
%\end{figure}

\subsection{Evaluation Metrics}

%The final measure of the quality of any haptic rendering algorithm should be captured by its realism performance based on human evaluation through a user study. However, such evaluations depend not only on the quality of the rendering algorithm but also the capability of the hardware being used. 
%In order to focus only on evaluating the performance of our model,
To evaluate the performance of our model, we directly compare the original signal with the output of our model for texture signal generation. Here, the term texture signal generation refers to predicting the induced acceleration due to the user's action with a material (i.e. their force and speed) during a tool-mediated interaction. As previous evidence suggests that humans' texture sensation is insensitive to phase shifts \cite{phasenotmatter}, we only take into account the magnitude of the discrete Fourier transform of this generated acceleration. Furthermore, as the variation of the user's force and speed during interaction results in a non-stationary signal, we compare the spectral distance of the two signals in short time windows and average the results over a shifting window. In our evaluation, we used a window size of 0.1 sec (1000 data-points) with a step size of 100 data-points and a Euclidean distance measure. Furthermore, we only calculated the spectral distance for frequencies up to a 1000 Hz as the original acceleration signal in the dataset was low-pass filtered at that frequency and it corresponds to human skin vibrotactile stimuli frequency threshold \cite{goff}.

%Whether other distance metrics in the spectral domain can be better choices than the euclidean distance in our evaluations is a research question that requires extensive human studies to be conducted in the future. However, it should be noted that our method is able to adapt to any new metric by simply learning to optimize that new loss function as its cost function during training. 
 
Finally, as a qualitative metric, we gain insight regarding the texture encoder in our generative model by visualizing the high-dimensional texture representation vector in our model (the output of texture encoder in Fig. \ref{fig:Architecture}) in a 2D space using a dimension reduction technique called t-Distributed Stochastic Neighbor Embedding (t-SNE) \cite{TSNE}. 

\subsection{Baseline}
We compare our model to the piece-wise autoregressive model used in \cite{Culbertson2014OneHD} which is a state-of-art method for vibration-based texture rendering. For a direct comparison to our model, we refit only the training and validation sections of the data into a piece-wise AR model and use it to synthesize the acceleration for the test set. This approach fits a separate model for each of the materials and requires manual labeling to match a new instance of a material with existing models.

\subsection{Generalization to New Interactions}

\begin{figure*}[t]
\begin{center}
%\fbox{\rule{0pt}{2in} \rule{0.9\linewidth}{0pt}}
    \includegraphics[width=\textwidth]{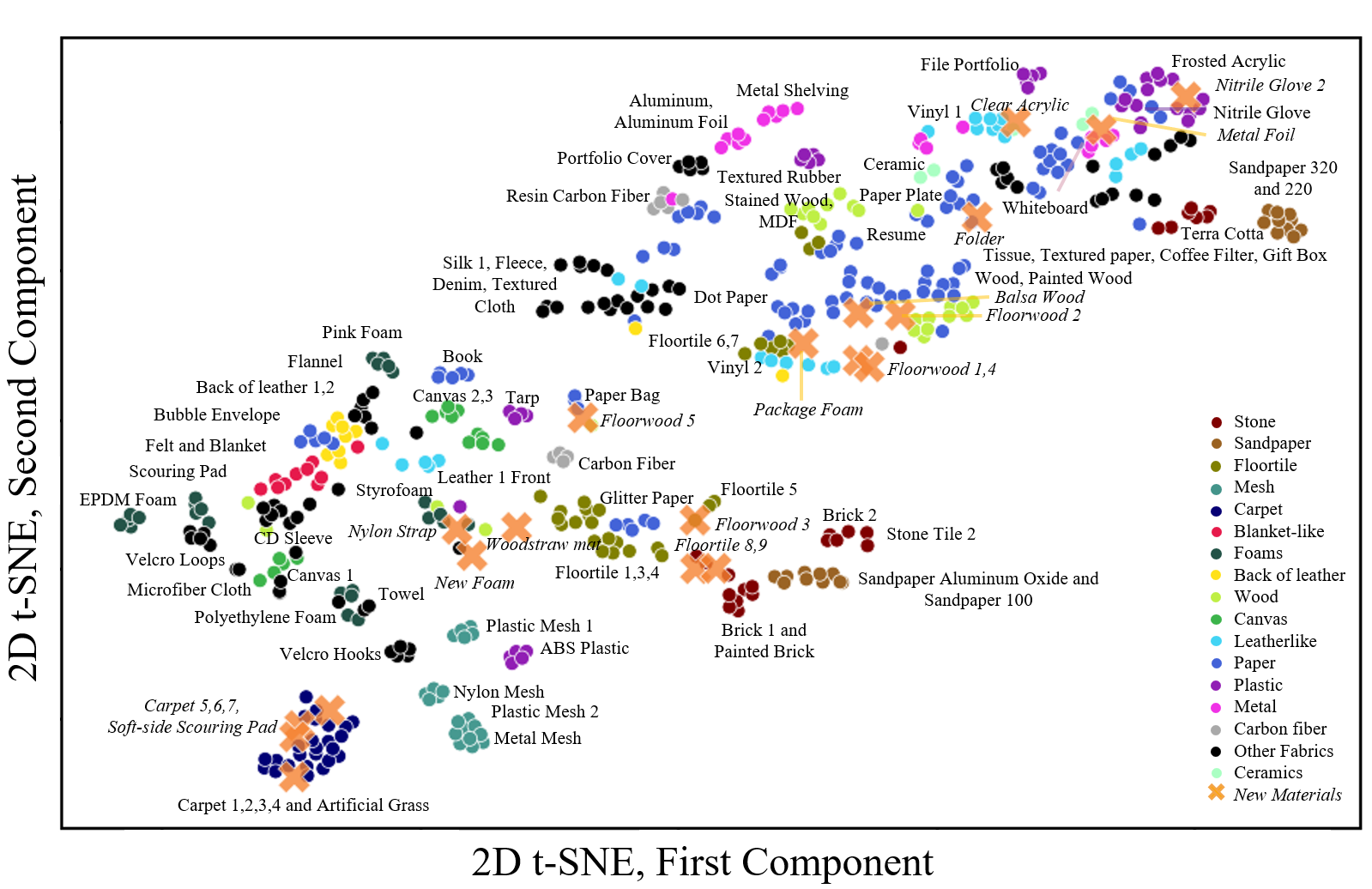} % it was 0.998
\end{center}
\caption{t-SNE of the learned latent representation for GelSight images in the training, validation, and test set and the encoding for new materials not in the training set. This 2D visualization of the high dimensional representation of the materials in the training set suggests that the neural network has learned to associate materials that feel and look similar in its latent space. Furthermore, the crosses showing the new materials not in the training set provide preliminary evidence that our approach has the potential to generalize to new materials (in italics) only using their GelSight image specially for those with distinct textural features (e.g. carpets).}
\label{fig:tsne}
\end{figure*} 

\label{sec:Unified}
For a direct comparison to the baseline, we first train one model per material by only feeding the action representation vector into the acceleration predictor module and removing the texture representation vector from the structure (since it is only one texture per model). Fig. \ref{fig:Shortterm} (top) shows the comparison of this trained model with the piece-wise AR model by showing the difference between the Euclidean errors of these two models on test set. Due to the randomness associated with the output of the piece-wise AR model, it was run 10 times on the test set and the average of the difference between all these runs was calculated. Our model has a lower average loss for 75 out of 93 materials compared to the baseline model. Error bars indicate 25 and 75 percent quartiles. This method significantly outperforms the baseline on 65 materials using Welch's t-test (p$<$0.05).

% The vertical version 

%\begin{figure}[t]
%\begin{center}
%\fbox{\rule{0pt}{2in} \rule{0.9\linewidth}{0pt}},angle=90
%\includegraphics[width=0.835\textwidth, angle = 270]{fig/Sep10ABBoth.png}
 %  \caption{Comparison of the AR model with the NN based model for (top) one model per material,(middle) unified model for all materials and (bottom) unified model for all materials on the temporal constructed signal using GLA on a local horizon (0.1 sec). This histogram represents the difference between the loss of  the ARMA prediction with that of a NN prediction. The error bars indicate the 25 and 75 percent quartiles. The loss for each model is defined as the L2 distance between the local short term DFT of its prediction and the corresponding spectral content of the ground truth acceleration.}
%   \wenzhen{I think in the caption, you could make the result more explicit. For example, explicitly say that 'A positive value in the bar plot means our method outperforms AR model'}
%   \negin{
 %  Note to self to do: change the font of the title to Times (major)}
%\label{fig:Shortterm}
%\end{center}
%\end{figure}

Afterwards, we trained a joint model for all the materials by adding the GelSight image as an input to the model. To assess the generalization capabilities of this unified model only in its capabilities to generalize to new actions, we keep the GelSight image input the same as the ones included in the training set and use the force and speed in the test set as input. This would be similar to a scenario where the material's label is known and one is trying to render it given a new action. The average results of the comparison of this model to the baseline as well as the 25 and 75 percent quartiles are shown in Fig. \ref{fig:Shortterm} (middle). This unified model has a lower average loss for 82 out of 93 materials compared to the baseline which is an improvement on the model in Fig. \ref{fig:Shortterm} (top). This method significantly outperforms the baseline on 78 materials using Welch's t-test (p$<$0.05).

Looking at the t-SNE visualization of the representation vector learned by the material encoder provides insight into the source of such improvement. This method visualizes high-dimensional vectors (here of the size 256) in a lower-dimensional space (here 2D). Fig. \ref{fig:tsne} shows that our texture encoder has learned to place the materials that feel similar closer to each other. For example, our encoder has created approximate clusters for all carpets as well as artificial grass (which feels like carpet), meshes, similar floortiles, stones, and similar sandpapers. %It also has placed meshes with square grids (plastic mesh 2, metal mesh) closer to each other than oval and circular shaped grids. 
Our hypothesis is that a unified model enables our network to share data between similar materials and cover a wider range of force and speed. Thereby, the unified model achieves an improved generalization performance.

\subsection{Generalization to New GelSight Images and Actions}
Fig. \ref{fig:Shortterm} (bottom) shows the performance of our trained unified model on generalizing to new GelSight images (of the training material) as well as new actions.
The baseline model requires manual labeling of the materials, so we report comparisons with respect to the baseline model's performance on new actions only. Our model has a lower average loss than baseline for 77 out of 93 models, without %while not requiring %any form of 
manual labeling. It also significantly outperforms the baseline on 71 materials using Welch's t-test (p$<$0.05).

%\begin{figure*}[ht]
%\begin{center}
%\fbox{\rule{0pt}{2in} \rule{0.9\linewidth}{0pt}}
%\includegraphics[width=1\linewidth]{fig/Sep10NewImage.png}
%\end{center}
%   \caption{Comparison of the AR model for new actions with that of the NN model for new actions as well as new images. To increase figure clarity, the FFT plots are only being shown up to the 1000 Hzs.}
%\label{fig:newboth}
%\end{figure*}

\subsection{Performance Evaluation of Constructed Temporal Signal}

\begin{figure*}%[ht]
\begin{center}
%\fbox{\rule{0pt}{2in} \rule{0.9\linewidth}{0pt}}
\includegraphics[width=0.96\linewidth]{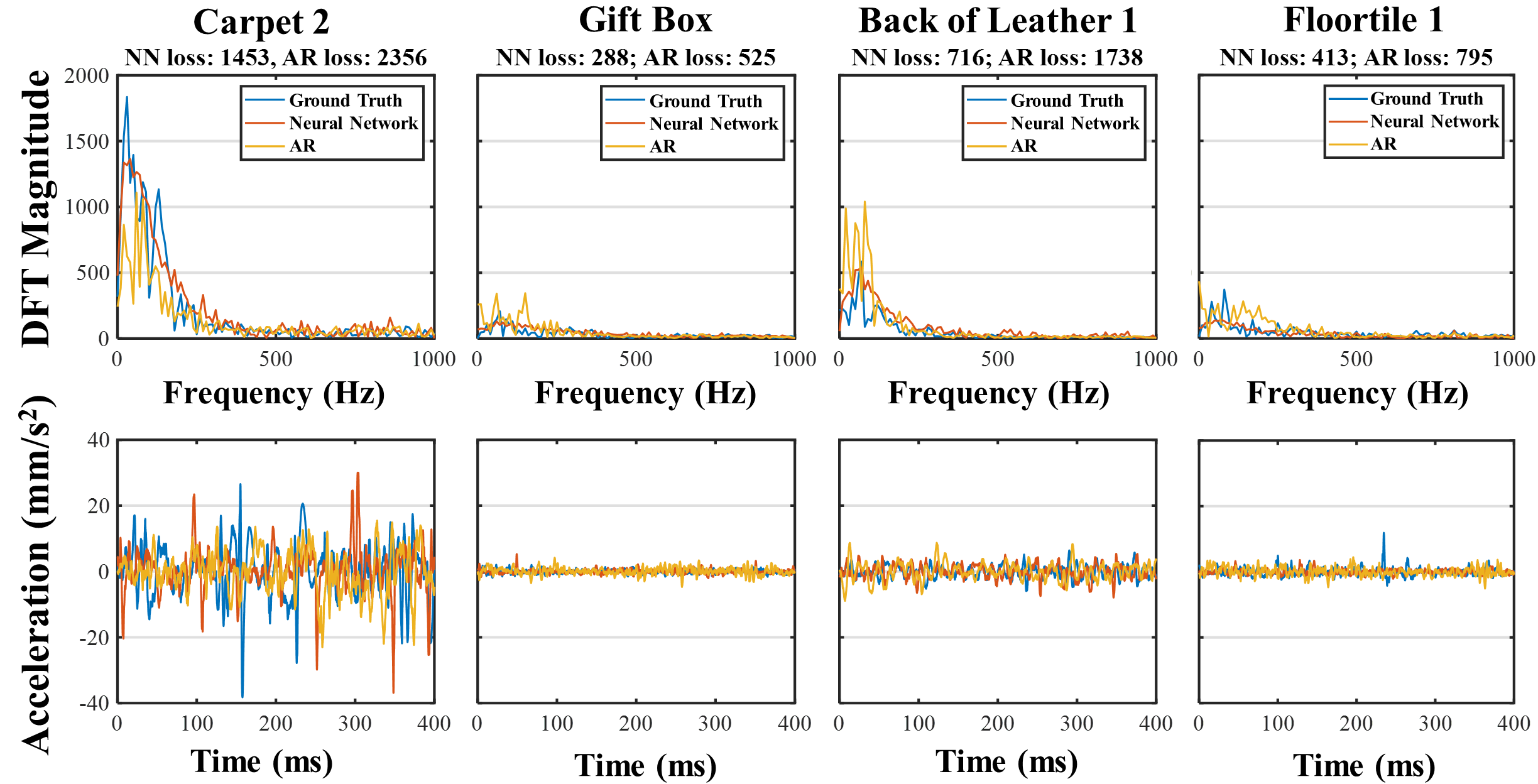}
\end{center}
   \caption{Comparison of the generated vibration by our model to ground truth and that of the Piece-wise AR model for four sample materials on test set.}
\label{fig:ComparePARMA}
\end{figure*}

As described in detail in Section \ref{HapticRendering}, the output predictions of our neural network structure on short horizons should be combined to create a temporal signal for haptic rendering. We found the Euclidean loss evaluation performance of such a stitched signal using GLA on the short horizon output of our unified model (our best performing model described in Section \ref{sec:Unified}) to have a lower average loss on 86.02\% of the materials compared to the baseline. This value is close to that of the non-stitched signal (88.17\%) suggesting that the temporal signal has kept its local spectral properties even after being stitched into a long horizon, which is desirable. Using Welch's t-test (p$<$0.05), this method significantly outperforms the baseline on 52 and underperforms on only 2 out of 93 materials. Due to space constraints, the performance bar plots are not shown.

%Using the proof-of-concept real-time stitching approach explained in Section \ref{HapticRendering}, we found the model to still outperform the baseline on 55.9\% of the materials. This can be even further improved by using existing optimization-based online phase retrieval and signal reconstruction algorithms in the future.% in this section.

To provide context for these numerical values, Fig. \ref{fig:ComparePARMA} shows 4 sample signals in the temporal as well as the spectral domains for the first 0.1 sec of the temporal signal constructed using GLA. It should be noted that for some materials, even though our method outperforms the previous method, on an absolute scale the prediction can still be far from the original signal. Floortile 1 in Fig. \ref{fig:ComparePARMA} captures such incident. A possible cause for this can be the relatively small dataset. Furthermore, other factors such as the grip force of the user while holding the data collection pen can affect the output and are not accounted for in our model. We believe collecting a larger automated dataset using a robot can further improve performance in the future. 

% It can be observed that different materials display different outputs  due to interactions both with respect to the magnitude of the FFT as well as the shape and magnitude of their acceleration.( I don't think this is 100% correct to say since the force and speed also varies between the samples I wonder if there's a way to say this but acknowledge that the force and speed also differs among them 

\subsection{Generalization to New Materials}

We provide preliminary evidence regarding our model's capability for generalizing to materials not in its training set by looking at the placement of the latent representation of these materials in the t-SNE space. The orange crosses in Fig. \ref{fig:tsne} represent 20 new materials. Materials with carpet-like textures were placed near other carpets. The new floortiles as well as Floorwood 3 were also placed close to the other floortiles. Metal foil, which had a very similar look and feel to whiteboard, has been placed close to it. Our method has also placed Balsa wood, Floorwood 2, and Floorwood 1,4 in the same region as wood and painted wood. Clear acrylic is also placed near other smooth plastics such as file portfolio and vinyl. Our method had difficulty generalizing for materials with significantly different texture than those in the dataset, such as nylon strap, woodstraw mat, a uniquely patterned foam, and package foam. 

\section{Conclusion and Future Work}
\label{sec:Conclusion}
In this paper, we model the vibratory signal that different textures induce in a hand-held tool moved over a surface. The action-conditional model takes as input a GelSight image of the texture as well as the force and speed the user applies to the texture, and outputs the desired acceleration for haptic texture generation. Furthermore, we augmented the HaTT dataset with GelSight images. In the future, we will evaluate our model during haptic rendering in a human user study and collect data from a wider range of materials with an autonomous robot to reduce human effort. We also plan to investigate the effects of replacing GelSight images with RGB images. We can also investigate the effects of using other loss measures for evaluating the performance of a model for haptic texture rendering.

%\footnote{Acknowledgement: we sincerely thank Katherine Kuchenbecker, Yasemin Vardar, Shaoxiong Wang, and Heather Culbertson for their help.}

%%%%%%%%%%%%%%%%%%%%%%%%%%%%%%%%%%%%%%%%%%%%%%%%%%%%%%%%%%%%%%%%%%%%%%%%

\bibliographystyle{IEEEtranN}
\newpage
\bibliography{Hapticbib}

\end{document}